\documentclass[]{spie}  

\usepackage{amsmath,amsfonts,amssymb}
\usepackage{graphicx}
\usepackage{float}  
\usepackage[colorlinks=true, allcolors=blue]{hyperref}
\usepackage{setspace}  
\usepackage{xcolor}
\usepackage[dvipsnames]{xcolor}
\usepackage{subcaption}
\usepackage{pifont}

\usepackage{times}

\usepackage{ulem}

\makeatletter
\newcommand{\rom}[1]{\expandafter\@slowromancap\romannumeral\number#1@}
\makeatother

\title{VAMOS-OCTA: Vessel-Aware Multi-Axis Orthogonal Supervision for Inpainting Motion-Corrupted OCT Angiography Volumes}

\author[a]{Nick DiSanto}
\author[a]{Ehsan Khodapanah Aghdam}
\author[b]{Han Liu}
\author[c]{Jacob Watson}
\author[c]{Yuankai K. Tao}
\author[a]{Hao Li}
\author[a]{Ipek Oguz}

\affil[a]{Department of Computer Science, Vanderbilt University}
\affil[b]{Digital Technology and Innovation, Siemens Healthineers}
\affil[c]{Department of Biomedical Engineering, Vanderbilt University}

\authorinfo{Further author information: (Send correspondence to Nick DiSanto)\\Nick DiSanto: E-mail: nicolas.c.disanto@vanderbilt.edu}

\begin{document}









\maketitle

\begin{abstract}
Handheld Optical Coherence Tomography Angiography (OCTA) enables noninvasive retinal imaging in uncooperative or pediatric subjects, but is highly susceptible to motion artifacts that severely degrade volumetric image quality. Sudden motion during 3D acquisition can lead to unsampled retinal regions across entire B-scans (cross-sectional slices), resulting in blank bands in \textit{en face} projections.
We propose \textbf{VAMOS-OCTA}, a deep learning framework for inpainting motion-corrupted B-scans using vessel-aware multi-axis supervision. We employ a 2.5D U-Net architecture that takes a stack of neighboring B-scans as input to reconstruct a corrupted center B-scan, guided by a novel \textbf{Vessel-Aware Multi-Axis Orthogonal Supervision (VAMOS)} loss. This loss combines vessel-weighted intensity reconstruction with axial and lateral projection consistency, encouraging vascular continuity in native B-scans and across orthogonal planes. Unlike prior work that focuses primarily on restoring the \textit{en face} MIP, VAMOS-OCTA jointly enhances both cross-sectional B-scan sharpness and volumetric projection accuracy, even under severe motion corruptions.
We trained our model on both synthetic and real-world corrupted volumes and evaluated its performance using both perceptual quality and pixel-wise accuracy metrics. VAMOS-OCTA consistently outperforms prior methods, producing reconstructions with sharp capillaries, restored vessel continuity, and clean \textit{en face} projections. These results demonstrate that multi-axis supervision offers a powerful constraint for restoring motion-degraded 3D OCTA data. Our source code is available at \url{https://github.com/MedICL-VU/VAMOS-OCTA}.
\end{abstract}


\section{Introduction}

Optical Coherence Tomography Angiography (OCTA) provides high-resolution volumetric images of microvascular networks in vivo, offering a non-invasive alternative to traditional dye-based angiography. However, handheld OCTA imaging of awake subjects is particularly susceptible to motion artifacts caused by eye movements and operator-induced shifts \cite{anvari2021artifacts}. Since volumes are acquired slice-by-slice, even brief motion events can corrupt entire B-scans or small blocks of contiguous B-scans, resulting in blank bands in \textit{en face} views such as the Maximum Intensity Projection (MIP).

While inter-frame registration \cite{yang2017handheld} and progressive sampling \cite{lebed2010rapid} methods have been proposed for motion correction, alignment-based approaches fail to recover fully unsampled retinal regions caused by bulk motion events. This limitation has motivated post-acquisition inpainting methods that attempt to reconstruct vessels from adjacent B-scans \cite{lin2024deep, li2021deep}. However, existing restoration approaches exhibit distinct and recurring failure modes. For example, 2D generative models, while visually compelling in isolation, can produce spatially misaligned reconstructions when applied independently across slices without explicit 3D constraints \cite{zhu2023make}. Additionally, many OCTA-specific restoration methods emphasize projection-domain improvements \cite{cao2023two, wang2023signal, kadomoto2020enhanced}, which fail to enforce local B-scan fidelity and can obscure slice-wise inconsistencies.

One post-acquisition inpainting approach, SOAD (Self-supervised OCTA Denoising) \cite{li2024self}, introduced blind-slice prediction using a vessel-weighted loss that emphasizes high-intensity angiographic signal. While this strategy enables self-supervised learning without requiring perfect ground-truth or fully-corrupted volumes, self-supervised models that use intensity-based losses have shown to produce over-smoothed reconstructions and blurry results that lose fine details \cite{zhang2017beyond, lin2025speckle2self}. Therefore, emphasizing vessel intensity alone is insufficient in practice, motivating a method with stronger constraints that optimize perceptual and feature-based losses for both individual B-scans and \textit{en face} projections.

In this work, we introduce \textbf{VAMOS-OCTA} (Vessel-Aware Multi-Axis Orthogonal Supervision for Inpainting Motion-Corrupted OCT Angiography Volumes), a deep learning framework designed to restore corrupted B-scans in handheld OCTA volumes while restoring vessels across orthogonal axes. Our main contributions are as follows:
\begin{enumerate}
    \item A composite vessel-aware loss function, termed the \textbf{VAMOS loss}, which combines vessel-weighted reconstruction with anatomically aligned projection constraints to prioritize accurate vessel restoration.
    \item A dual-axis projection supervision strategy for OCTA inpainting that leverages both axial and lateral projections to enforce consistency across orthogonal planes.
    \item Joint enhancement of 2D and 3D representations, yielding smooth \textit{en face} projections and perceptually sharp B-scans, evaluated using both quality  and accuracy metrics.
\end{enumerate}

\begin{figure}[b]
    \centering
    \includegraphics[width=1.0\textwidth]{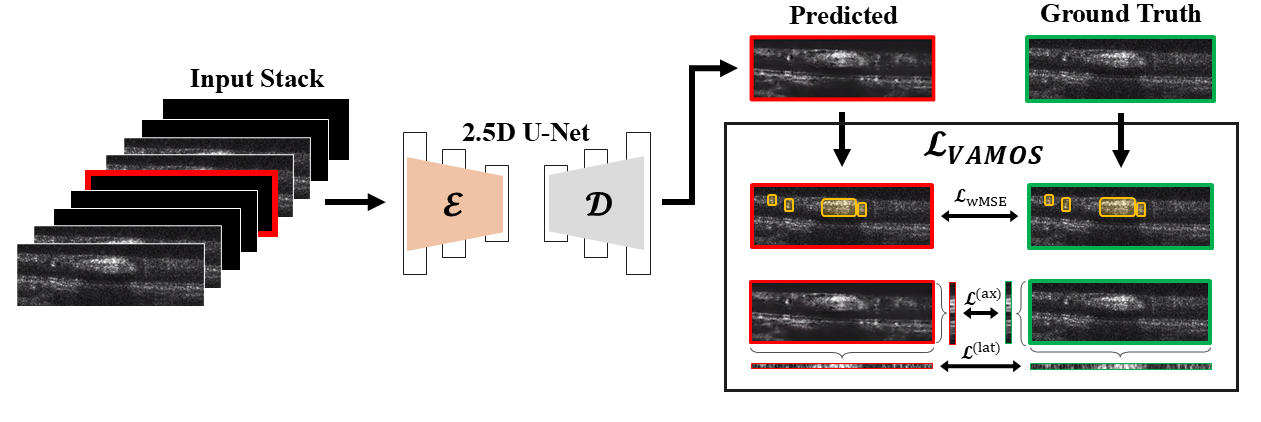}
    \caption{Overview of the VAMOS-OCTA framework. A 2.5D U-Net reconstructs a central target B-scan from a stack of neighboring B-scans. Training is guided by the proposed VAMOS loss, which combines vessel-weighted reconstruction with multi-axis orthogonal supervision. Axial and lateral projection constraints are enforced by collapsing each B-scan along depth and lateral dimensions, respectively, to form 1D maximum and average intensity profiles.}
    \label{fig:vamos_framework}
\end{figure}

\noindent
Together, these contributions establish a robust and anatomically aware inpainting framework for restoring motion-corrupted OCTA volumes, supporting reliable downstream visualization and clinical interpretation. Our overall framework is summarized in Figure~\ref{fig:vamos_framework}.

\section{Methods}

\subsection{Data}

\subsubsection{Acquisition and Preprocessing}

We use 7 high-resolution volumetric retinal scans acquired from awake patients using a custom handheld multimodal OCT/OCTA probe that allows for simultaneous \textit{en face} guidance and co-registered OCT acquisition \cite{watson2025point}. Volumes were initially acquired as structural OCT with repeated B-scans per position. Axial motion correction was applied using subpixel row-wise registration, after which OCTA volumes were generated using an SVD-based decorrelation algorithm that derives angiographic contrast from temporal intensity variations across repeated B-scans. The resulting OCTA volumes are 3D stacks of 1000 B-scans, each of size $400 \times 145$ \textit{(lateral $\times$ axial)} cropped to the retinal region of interest. The dataset was split 5/1/1 for training/validation/testing, rotated across 7 cross-validation folds.

\subsubsection{Synthetic Motion Corruption Simulation}
To obtain paired corrupted and ground-truth data for training and evaluation, we adopt a synthetic motion corruption strategy designed to simulate bulk motion events commonly observed in handheld OCTA acquisition. Specifically, for each target B-scan to be reconstructed, we remove the center slice and additionally drop a random number of neighboring B-scans to simulate contiguous unsampled retinal regions arising from a bulk motion event. The size of each corrupted block is sampled from a geometric distribution with parameter $p = 0.4$, truncated to a maximum length of 6 slices. This distribution biases the corruption process toward short-to-moderate dropout lengths, reflecting empirical observations of handheld motion patterns \cite{hormel2021artifacts, ni2024panretinal}. During training, this corruption process is applied dynamically for each target B-scan at every epoch, ensuring that the network observes a diverse range of motion patterns and corruption configurations over time. This prevents overfitting to fixed corruption patterns and encourages robustness to variable corruption severity. In contrast, fixed corruption masks are used during validation and testing to ensure consistent and fair comparison across methods.

\subsection{Validity-Aware 2.5D U-Net}

To support volumetric inpainting under dynamic motion-induced corruptions, we modify the standard U-Net \cite{ronneberger2015u} to reconstruct a target B-scan based on its spatial context. More specifically, we employ a 2.5D architecture that accepts an input stack of neighboring B-scans, where some slices may have unsampled retinal regions due to simulated motion artifacts, and inpaints the B-scan at the center of the stack. The network's input is a tensor of shape $(B, S, H, W)$, where $B$ is the batch size, $S$ is the number of B-scans in the input stack (an odd number, set to $S=9$ in our experiments), and $(H, W)$ are the axial and lateral dimensions of each B-scan. The stochastic corruptions applied to each stack during training allow the model to learn from partially missing context by relying on the valid, uncorrupted neighbors.

The network does not require a fully intact input stack or an explicit validity mask. Instead, it learns to selectively rely on valid neighboring B-scans while down-weighting corrupted context. Through exposure to diverse corruption patterns during training, the network learns to perform robust inpainting under partial contextual information.

\subsection{VAMOS Loss: Vessel-Aware Multi-Axis Orthogonal Supervision}

The core of our framework is the VAMOS loss, which is designed to jointly preserve local vessels within individual B-scans and global anatomical consistency across volumetric projections. While SOAD's weighted reconstruction loss attempts to emphasize accuracy across vessels, it provides limited control over B-scan texture and \textit{en face} projections under severe motion corruption. To address this limitation, we augment vessel-weighted reconstruction with projection-based supervision that constrains continuity across orthogonal planes. Formally, the VAMOS loss is defined as:
\begin{equation}
\mathcal{L}_{\text{VAMOS}} = \mathcal{L}_{\text{wMSE}} + \lambda_{\text{proj}}
\left(
\mathcal{L}^{(ax)}_{\text{MIP}} + \mathcal{L}^{(lat)}_{\text{MIP}} + \mathcal{L}^{(ax)}_{\text{AIP}} + \mathcal{L}^{(lat)}_{\text{AIP}}
\right)
\end{equation}


\noindent
This is composed of a vessel-weighted mean squared error (wMSE) reconstruction loss, along with L1 projection losses for the Maximum Intensity Projection (MIP) and Average Intensity Projection (AIP) distributions along axial and lateral axes. These losses will be detailed in the following subsections. In all experiments, we set $\lambda_{\text{proj}}=3$, giving greater emphasis to projection consistency relative to pixel-wise reconstruction.

\subsubsection{Vessel-Weighted MSE}
To match general intensities while prioritizing bright vessel pixels, we use a vessel-weighted L2 loss as proposed by SOAD \cite{li2024self}. Let $Y(x,z)$ and $\hat{Y}(x,z)$ denote the ground-truth and reconstructed OCTA B-scans, respectively, where $x$ indexes the lateral dimension and $z$ indexes the axial (depth) dimension. The vessel-weighted reconstruction loss is defined as:
\begin{equation}
\begin{aligned}
\mathcal{L}_{\text{wMSE}} &= \frac{1}{N} \sum_{x,z} w(x,z) \left[ \hat{Y}(x,z) - Y(x,z) \right]^2,
\hspace{0.7em}\text{with}\hspace{0.7em}
w(x,z) =
\underbrace{ \alpha_w \left[ \hat{Y}(x,z) \right]^{\gamma_w} }_{\text{prediction-based}} +\;
\underbrace{ \left[ Y(x,z) \vphantom{ \hat{Y}(x,z) } \right]^{\gamma_w} }_{\text{target-based}} +\; c
\end{aligned}
\end{equation}

\noindent
Here, the spatially varying weight $w(x,z) \ge 1$ amplifies loss contributions via a prediction-based term discouraging hallucinated bright spots, a target-based term emphasizing true vessels, and a sublinear exponent moderating outliers. Following SOAD, we set $\alpha_w = 100$, $\gamma_w = \tfrac{1}{3}$, and $c = 0.5$.

\subsubsection{Multi-Axis Projection Losses}
To regularize the vessel distribution in 3D, we introduce maximum and average projection losses over two orthogonal axes: \textbf{axial} and \textbf{lateral}.
\begin{equation}
\begin{aligned}
\mathcal{L}^{\text{(ax)}}_{\text{MIP}} &= \frac{1}{W} \sum_{x} \left| \max_{z} \hat{Y}(x, z) - \max_{z} Y(x, z) \right|, \quad
\mathcal{L}^{\text{(ax)}}_{\text{AIP}} = \frac{1}{W} \sum_{x} \left| \frac{1}{H} \sum_{z} \hat{Y}(x, z) - \frac{1}{H} \sum_{z} Y(x, z) \right| \\
\mathcal{L}^{\text{(lat)}}_{\text{MIP}} &= \frac{1}{H} \sum_{z} \left| \max_{x} \hat{Y}(x, z) - \max_{x} Y(x, z) \right|, \quad
\mathcal{L}^{\text{(lat)}}_{\text{AIP}} = \frac{1}{H} \sum_{z} \left| \frac{1}{W} \sum_{x} \hat{Y}(x, z) - \frac{1}{W} \sum_{x} Y(x, z) \right|
\end{aligned}
\end{equation}

\noindent
Axial projections constrain depth-integrated vessel continuity, directly influencing the appearance of \textit{en face} views, while lateral projections regularize the vessel distribution within each individual B-scan, mitigating banding artifacts. We employ both Maximum Intensity Projection (MIP) and Average Intensity Projection (AIP) losses along each axis. MIP losses emphasize peak vessel intensities, while AIP losses constrain the overall distribution of intensities, encouraging realistic vessel density.

By integrating these objectives, the VAMOS loss constrains both the intensity and spatial distribution of vessels across orthogonal axes. The vessel-weighted and projection-based terms work in tandem to enforce volumetric consistency on a slice-by-slice basis. Similar to recent cross-view projection refinement approaches \cite{radhakrishna2024spockmip}, this multi-axis supervision guides reconstruction toward volumetric consistency without requiring explicit 3D modeling.

 \begin{table}[b]
\centering
\caption{
Quantitative performance for B-scans and  MIPs.
\textbf{Bold} indicates best performance. Asterisks (*) indicate statistically significant improvements over SOAD ($p < 0.05$, paired $t$-test).}

\begin{subtable}[t]{\textwidth}
\centering
\resizebox{\textwidth}{!}{%
\begin{tabular}{|l|c|c|c|c|c|}
\hline
\textbf{Method} & \textbf{Gradient L1↓} & \textbf{LPIPS↓} & \textbf{Laplac Blur Diff↓} & \textbf{Sobel Edge Pres↑} & \textbf{PSNR↑} \\
\hline
Standard MSE & 0.045 ± 0.014 & 0.625 ± 0.199 & 0.023 ± 0.008 & 0.305 ± 0.220 & 27.036 ± 0.764 \\
SOAD Weighted MSE (wMSE)\cite{li2024self} & 0.044 ± 0.014 & 0.608 ± 0.202 & 0.023 ± 0.008 & 0.313 ± 0.219 & \textbf{27.084 ± 0.712} \\
wMSE + Axial Projection & 0.042 ± 0.013* & 0.565 ± 0.184* & 0.020 ± 0.007 & 0.360 ± 0.204* & 26.687 ± 0.739 \\
wMSE + Axial + Lateral Proj (\textbf{VAMOS-OCTA}) & \textbf{0.041 ± 0.013*} & \textbf{0.510 ± 0.162*} & \textbf{0.014 ± 0.005*} & \textbf{0.427 ± 0.182*} & 26.122 ± 0.709 \\
\hline
\end{tabular}
}
\caption{B-scan reconstruction performance using perceptual quality metrics: Gradient L1, Learned Perceptual Image Patch Similarity (LPIPS) \cite{zhang2018unreasonable}, Laplacian Blur Difference, Sobel Edge Preservation.}
\label{tab:table-bscan-stacked}
\end{subtable}

\begin{subtable}[t]{\textwidth}
\centering
\resizebox{\textwidth}{!}{%
\begin{tabular}{|l|c|c|c|c|c|}
\hline
\textbf{Method} & \textbf{L1↓} & \textbf{MIE↓} & \textbf{SSIM↑} & \textbf{NCC↑} & \textbf{PSNR↑} \\
\hline
Standard MSE & 0.029 ± 0.003 & 0.028 ± 0.003 & 0.758 ± 0.011 & 0.808 ± 0.008 & 22.550 ± 0.818 \\ 
SOAD Weighted MSE (wMSE)\cite{li2024self} & 0.028 ± 0.003 & 0.029 ± 0.003 & 0.763 ± 0.008 & 0.812 ± 0.006 & 22.672 ± 0.717 \\ 
wMSE + Axial Projection & 0.014 ± 0.002* & 0.005 ± 0.004* & 0.888 ± 0.008* & 0.914 ± 0.006* & \textbf{28.168 ± 1.044*} \\
wMSE + Axial + Lateral Proj (\textbf{VAMOS-OCTA}) & \textbf{0.013 ± 0.002*} & \textbf{0.003 ± 0.003*} & \textbf{0.895 ± 0.010*} & \textbf{0.931 ± 0.007*} & 27.772 ± 0.865* \\
\hline
\end{tabular}
}
\caption{MIP reconstruction performance using pixel-wise accuracy metrics: L1, Mean Intensity Error (MIE), Structural Similarity Index (SSIM) \cite{wang2004image}, Normalized Cross-Correlation (NCC), Peak Signal-to-Noise Ratio (PSNR).}
\label{tab:table-mip-stacked}
\end{subtable}
\label{tab:table-mip-bscan-stacked}
\end{table}

\section{Results}

\subsection{Quantitative Performance}

We trained all models from scratch on an NVIDIA RTX 6000 Ada GPU (CUDA 12.4) using dynamic synthetic motion artifacts. During testing, fixed corruption masks were used and held constant across methods to ensure fair comparison.
Since pixel-wise metrics often favor blurring under uncertainty \cite{terven2025comprehensive}, we employ perceptual and edge-aware metrics for B-scan evaluations to capture sharpness and vessel detail. We report LPIPS \cite{zhang2018unreasonable} for perceptual similarity, Laplacian blur difference for sharpness, and a Sobel gradient edge preservation ratio to quantify vessel boundary retention (Table~\ref{tab:table-bscan-stacked}).
For \textit{en face} MIP reconstruction, we use the pixel-wise accuracy metrics L1, MIE, SSIM, NCC, and PSNR to evaluate depth-integrated volumetric consistency, ensuring projection-level anatomical accuracy (Table~\ref{tab:table-mip-stacked}).

\noindent
Our results show that SOAD encourages the preservation of large vessels but favors smoothness over fine detail, resulting in weaker perceptual and edge-aware performance. Adding axial projection supervision improves MIP quality, yielding notable gains in SSIM and NCC, but offers limited benefit to B-scan sharpness. In contrast, VAMOS-OCTA, which incorporates lateral projection supervision, achieves consistent improvements across both perceptual B-scan metrics and projection accuracy. This confirms that lateral constraints are necessary to suppress banding artifacts and maintain realistic vessel structure. Nearly all of these gains are statistically significant and consistent across all cross-validation folds, underscoring that multi-axis supervision robustly enhances quantitative performance without trade-offs.

\begin{figure}[t]
    \centering
    \includegraphics[width=1.0\textwidth]{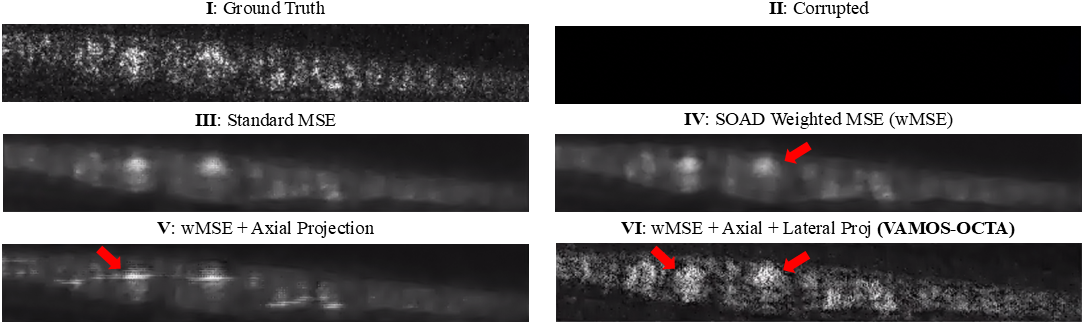}
    \caption{Reconstructed B-scan for each method. Arrows highlight areas of interest:~(\protect\rom{4}) oversmooths vessels and~(\protect\rom{5}) introduces horizontal banding, while VAMOS-OCTA (\protect\rom{6}) restores vessel contrast and spatial continuity.}
    \label{fig:figure-bscan-stacked}
\end{figure}
\begin{figure}[b]
    \centering
    \includegraphics[width=1.0\textwidth]{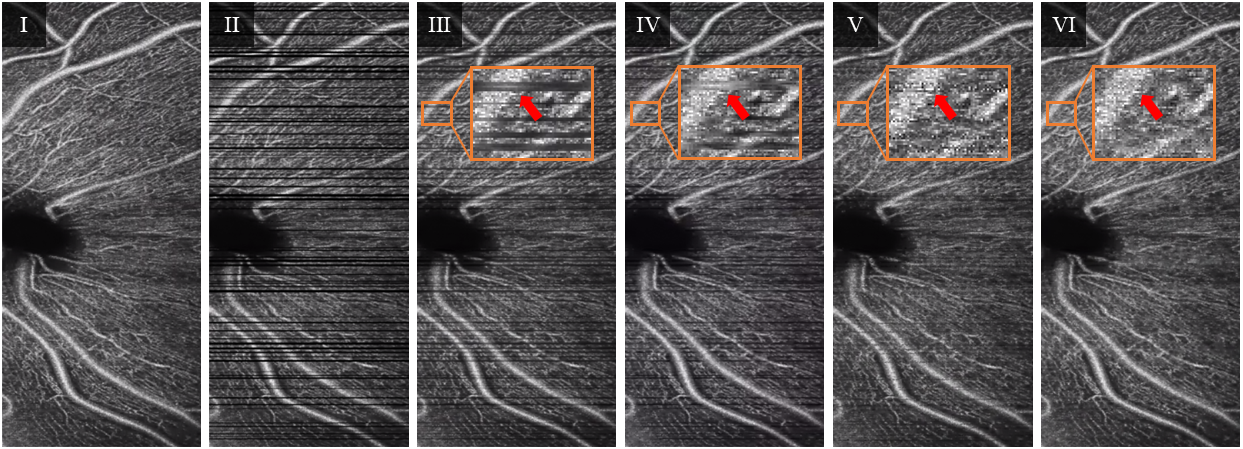}
    \caption{
    Depth-wise \textit{en face} MIP reconstructions. 
    (\textbf{I}) Ground Truth, (\textbf{II}) Corrupted, (\textbf{III}) Standard MSE, 
    (\textbf{IV}) SOAD Weighted MSE, (\textbf{V}) wMSE + Axial, 
    (\textbf{VI}) VAMOS-OCTA. Zoom panels highlight an area where
    (\textbf{III}), (\textbf{IV}), and (\textbf{V}) fail to remove artifacts, while VAMOS-OCTA (\textbf{VI}) restores a smooth and realistic projection with recovered vessels (arrows).}
    
    \label{fig:figure-mip}
\end{figure}

\subsection{Qualitative Performance}

\begin{figure}[t]
    \centering
    \begin{minipage}[t]{0.48\textwidth}
        \input{figure-real-motioncorrected-inpainting}
    \end{minipage}
    \hspace{0.8em}
    \begin{minipage}[t]{0.49\textwidth}
        \input{figure-performance-vs-corruption-severity}
    \end{minipage}
\end{figure}

Figure~\ref{fig:figure-bscan-stacked} presents qualitative reconstructions comparing individual B-scans and corresponding \textit{en face} MIPs produced by baseline MSE, SOAD, partial projection supervision, and the full VAMOS-OCTA model, each under severe motion-induced sampling gaps. Baseline MSE and SOAD reconstructions are over-smoothed in individual B-scans and produce fragmented vessels in the MIP. While introducing axial projection supervision partially mitigates these issues by improving MIP reconstruction, it aggregates information across depth and thus fails to fully constrain the vessels across both axial and lateral axes. This results in horizontal banding artifacts, which disrupt B-scan consistency.

In contrast, the full VAMOS-OCTA model, which incorporates both axial and lateral projection supervision, yields qualitatively consistent improvements across both B-scans and the \textit{en face} MIP. Lateral projection constraints explicitly regularize slice-wise vessel placement, suppressing banding artifacts and restoring continuity across neighboring B-scans. This results in sharp texture, realistic vessel branching, and microvascular continuity that is lost in competing methods. VAMOS-OCTA's inpainting quality is also maintained across slices, enabling \textit{en face} MIPs (Figure~\ref{fig:figure-mip}) that are structurally coherent and free of residual dropout bands.

These qualitative observations directly align with the quantitative trends reported in Table~\ref{tab:table-mip-bscan-stacked}. While axial projection supervision primarily improves projection-domain metrics, multi-axis supervision is necessary to jointly recover perceptually sharp B-scans and volumetric structure. We further evaluate robustness by applying VAMOS-OCTA to real-world motion-corrupted volumes (Figure~\ref{fig:figure-real-motioncorrected-inpainting}) and volumes with increasingly large contiguous unsampled regions (Figure~\ref{fig:performance-vs-corrupted-severity}), both of which maintain strong performance. These improvements are also achieved without explicit 3D modeling, demonstrating that orthogonal projection constraints provide an effective mechanism for enforcing volumetric consistency in 2.5D OCTA inpainting.

\section{Discussion and Conclusion}
In this work, we introduced \textbf{VAMOS-OCTA}, a multi-axis orthogonal supervision framework for inpainting motion-corrupted OCTA volumes acquired with handheld imaging systems. By augmenting vessel-weighted reconstruction with axial and lateral projection constraints, VAMOS-OCTA produces perceptually sharp B-scans and structurally coherent \textit{en face} MIPs under severe motion-induced sampling gaps.

A central insight of this work is that explicitly constraining orthogonal projections provides a more effective form of volumetric regularization than intensity-based reconstruction alone. While vessel-weighted losses broadly encourage recovery of high-intensity regions, they offer limited control over the structure of reconstructed vessels. Incorporating projection-based supervision allows the model to constrain how vessels aggregate across depth and between slices, discouraging configurations that may satisfy local pixel-wise objectives but violate global continuity. The complementary roles of axial and lateral projections are critical: axial supervision improves depth-integrated vessel visibility, while lateral supervision improves inter-slice structure and suppresses banding artifacts.

We also demonstrate that volumetric consistency can emerge from strong 2D constraints applied per-B-scan, suggesting that it need not arise exclusively from explicit 3D architectures. Instead, carefully designed geometric constraints imposed at the loss level can induce consistent 3D structure even when operating on 2D or 2.5D representations. This aligns with recent work that improves 3D consistency via iterative slice generation \cite{jeong2023generating} and multi-view refinement \cite{shi2025imfine}. More broadly, our results contribute to a growing body of work on inpainting and restoration in volumetric medical imaging \cite{santos2025role, wang2023self, zhang2020robust}, reinforcing the importance of spatial context and cross-slice supervision. This work establishes orthogonal projection constraints as an effective strategy for volumetric OCTA restoration, offering a generalizable framework for addressing motion artifacts in other slice-based 3D imaging modalities.

\acknowledgments 
This work was supported, in part, by NIH grant R01-EY033969 and the Vanderbilt Advanced Computing Center for Research and Education.

\bibliographystyle{spiebib}
\bibliography{report}

@inproceedings{ronneberger2015u,
    title={U-net: Convolutional networks for biomedical image segmentation},
    author={Ronneberger, Olaf and Fischer, Philipp and Brox, Thomas},
    booktitle={International Conference on Medical image computing and computer-assisted intervention},
    pages={234--241},
    year={2015},
    organization={Springer}
}

@article{anvari2021artifacts,
    title={Artifacts in optical coherence tomography angiography},
    author={Anvari, Pasha and Ashrafkhorasani, Maryam and Habibi, Abbas and Falavarjani, Khalil Ghasemi},
    journal={Journal of Ophthalmic \& Vision Research},
    volume={16},
    number={2},
    pages={271},
    year={2021}
}

@article{yang2017handheld,
    title={Handheld optical coherence tomography angiography},
    author={Yang, Jianlong and Liu, Liang and Campbell, J Peter and Huang, David and Liu, Gangjun},
    journal={Biomedical optics express},
    volume={8},
    number={4},
    pages={2287--2300},
    year={2017},
    publisher={Optical Society of America}
}

@inproceedings{li2024self,
    title={Self-supervised Denoising and Bulk Motion Artifact Removal of 3D Optical Coherence Tomography Angiography of Awake Brain},
    author={Li, Zhenghong and Ren, Jiaxiang and Zou, Zhilin and Garigapati, Kalyan and Du, Congwu and Pan, Yingtian and Ling, Haibin},
    booktitle={International Conference on Medical Image Computing and Computer-Assisted Intervention},
    pages={601--611},
    year={2024},
    organization={Springer}
}

@article{radhakrishna2024spockmip,
    title={Spockmip: Segmentation of vessels in mras with enhanced continuity using maximum intensity projection as loss},
    author={Radhakrishna, Chethan and Chintalapati, Karthikesh Varma and Kumar, Sri Chandana Hudukula Ram and Sutrave, Raviteja and Mattern, Hendrik and Speck, Oliver and N{\"u}rnberger, Andreas and Chatterjee, Soumick},
    journal={arXiv preprint arXiv:2407.08655},
    year={2024}
}

@article{zhang2017beyond,
    title={Beyond a gaussian denoiser: Residual learning of deep cnn for image denoising},
    author={Zhang, Kai and Zuo, Wangmeng and Chen, Yunjin and Meng, Deyu and Zhang, Lei},
    journal={IEEE transactions on image processing},
    volume={26},
    number={7},
    pages={3142--3155},
    year={2017},
    publisher={IEEE}
}

@article{li2021deep,
  title={Deep-learning-based motion correction in optical coherence tomography angiography},
  author={Li, Ang and Du, Congwu and Pan, Yingtian},
  journal={Journal of biophotonics},
  volume={14},
  number={12},
  pages={e202100097},
  year={2021},
  publisher={Wiley Online Library}
}

@article{lebed2010rapid,
    title={Rapid volumetric OCT image acquisition using compressive sampling},
    author={Lebed, Evgeniy and Mackenzie, Paul J and Sarunic, Marinko V and Beg, Mirza Faisal},
    journal={Optics express},
    volume={18},
    number={20},
    pages={21003--21012},
    year={2010},
    publisher={Optical Society of America}
}

@inproceedings{watson2025point,
    title={Point-of-care widefield retinal OCTA mosaicking with handheld spectrally encoded coherence tomography and reflectometry},
    author={Watson, Jacob J and Xu, Rui and Hecht, Rachel and Tao, Yuankai},
    booktitle={Optical Coherence Tomography and Coherence Domain Optical Methods in Biomedicine XXIX},
    pages={PC1330507},
    year={2025},
    organization={SPIE}
}

@article{hormel2021artifacts,
    title={Artifacts and artifact removal in optical coherence tomographic angiography},
    author={Hormel, Tristan T and Huang, David and Jia, Yali},
    journal={Quantitative Imaging in Medicine and Surgery},
    volume={11},
    number={3},
    pages={1120},
    year={2021}
}

@article{ni2024panretinal,
  title={Panretinal handheld OCT angiography for pediatric retinal imaging},
  author={Ni, Shuibin and Liang, Guangru Ben and Ng, Ringo and Ostmo, Susan and Jia, Yali and Chiang, Michael F and Huang, David and Skalet, Alison H and Young, Benjamin K and Campbell, J Peter and others},
  journal={Biomedical Optics Express},
  volume={15},
  number={5},
  pages={3412--3424},
  year={2024},
  publisher={Optica Publishing Group}
}

@article{cao2023two,
  title={A two-stage framework for optical coherence tomography angiography image quality improvement},
  author={Cao, Juan and Xu, Zihao and Xu, Mengjia and Ma, Yuhui and Zhao, Yitian},
  journal={Frontiers in medicine},
  volume={10},
  pages={1061357},
  year={2023},
  publisher={Frontiers Media SA}
}

@article{wang2023signal,
  title={Signal attenuation-compensated projection-resolved OCT angiography},
  author={Wang, Jie and Hormel, Tristan T and Bailey, Steven T and Hwang, Thomas S and Huang, David and Jia, Yali},
  journal={Biomedical Optics Express},
  volume={14},
  number={5},
  pages={2040--2054},
  year={2023},
  publisher={Optica Publishing Group}
}

@inproceedings{jeong2023generating,
  title={Generating high-resolution 3d ct with 12-bit depth using a diffusion model with adjacent slice and intensity calibration network},
  author={Jeong, Jiheon and Kim, Ki Duk and Nam, Yujin and Cho, Kyungjin and Kang, Jiseon and Hong, Gil-Sun and Kim, Namkug},
  booktitle={International Conference on Medical Image Computing and Computer-Assisted Intervention},
  pages={366--375},
  year={2023},
  organization={Springer}
}

@inproceedings{zhu2023make,
  title={Make-a-volume: Leveraging latent diffusion models for cross-modality 3d brain mri synthesis},
  author={Zhu, Lingting and Xue, Zeyue and Jin, Zhenchao and Liu, Xian and He, Jingzhen and Liu, Ziwei and Yu, Lequan},
  booktitle={International Conference on Medical Image Computing and Computer-Assisted Intervention},
  pages={592--601},
  year={2023},
  organization={Springer}
}

@article{terven2025comprehensive,
  title={A comprehensive survey of loss functions and metrics in deep learning},
  author={Terven, Juan and Cordova-Esparza, Diana-Margarita and Romero-Gonz{\'a}lez, Julio-Alejandro and Ram{\'\i}rez-Pedraza, Alfonso and Ch{\'a}vez-Urbiola, EA},
  journal={Artificial Intelligence Review},
  volume={58},
  number={7},
  pages={195},
  year={2025},
  publisher={Springer}
}

@inproceedings{zhang2018unreasonable,
  title={The unreasonable effectiveness of deep features as a perceptual metric},
  author={Zhang, Richard and Isola, Phillip and Efros, Alexei A and Shechtman, Eli and Wang, Oliver},
  booktitle={Proceedings of the IEEE conference on computer vision and pattern recognition},
  pages={586--595},
  year={2018}
}

@inproceedings{shi2025imfine,
  title={Imfine: 3d inpainting via geometry-guided multi-view refinement},
  author={Shi, Zhihao and Huo, Dong and Zhou, Yuhongze and Min, Yan and Lu, Juwei and Zuo, Xinxin},
  booktitle={Proceedings of the Computer Vision and Pattern Recognition Conference},
  pages={26694--26703},
  year={2025}
}

@article{wang2004image,
  title={Image quality assessment: from error visibility to structural similarity},
  author={Wang, Zhou and Bovik, Alan C and Sheikh, Hamid R and Simoncelli, Eero P},
  journal={IEEE transactions on image processing},
  volume={13},
  number={4},
  pages={600--612},
  year={2004},
  publisher={IEEE}
}

@article{lin2025speckle2self,
  title={Speckle2Self: Learning Self-Supervised Despeckling with Attention Mechanism for SAR Images},
  author={Lin, Huiping and Su, Xin and Zeng, Zhiqiang and Xing, Cheng and Yin, Junjun},
  journal={Remote Sensing},
  volume={17},
  number={23},
  pages={3840},
  year={2025},
  publisher={MDPI}
}

@article{lin2024deep,
  title={Deep learning for motion artifact-suppressed octa image generation from both repeated and adjacent oct scans},
  author={Lin, Zhefan and Zhang, Qinqin and Lan, Gongpu and Xu, Jingjiang and Qin, Jia and An, Lin and Huang, Yanping},
  journal={Mathematics},
  volume={12},
  number={3},
  pages={446},
  year={2024},
  publisher={MDPI}
}

@article{kadomoto2020enhanced,
  title={Enhanced visualization of retinal microvasculature in optical coherence tomography angiography imaging via deep learning},
  author={Kadomoto, Shin and Uji, Akihito and Muraoka, Yuki and Akagi, Tadamichi and Tsujikawa, Akitaka},
  journal={Journal of Clinical Medicine},
  volume={9},
  number={5},
  pages={1322},
  year={2020},
  publisher={MDPI}
}

@article{santos2025role,
  title={The Role of Deep Learning in Medical Image Inpainting: A Systematic Review},
  author={Santos, Joana Cristo and Tom{\'a}s Pereira Alexandre, Hugo and Seoane Santos, Miriam and Henriques Abreu, Pedro},
  journal={ACM Transactions on Computing for Healthcare},
  volume={6},
  number={3},
  pages={1--24},
  year={2025},
  publisher={ACM New York, NY}
}

@article{wang2023self,
  title={Self-supervised CSF inpainting with synthetic atrophy for improved accuracy validation of cortical surface analyses},
  author={Wang, Jiacheng and Larson, Kathleen E and Oguz, Ipek},
  journal={arXiv preprint arXiv:2303.05777},
  year={2023}
}

@inproceedings{zhang2020robust,
  title={Robust multiple sclerosis lesion inpainting with edge prior},
  author={Zhang, Huahong and Bakshi, Rohit and Bagnato, Francesca and Oguz, Ipek},
  booktitle={International Workshop on Machine Learning in Medical Imaging},
  pages={120--129},
  year={2020},
  organization={Springer}
}

\end{document}